%% file: ms.tex
\documentclass{article}

\usepackage{microtype}
\usepackage{graphicx}
\usepackage{booktabs} %

\usepackage{hyperref}

\usepackage[accepted]{icml2025}

\usepackage{amsmath}
\usepackage{amssymb}
\usepackage{mathtools}
\usepackage{amsthm}

\usepackage[capitalize,noabbrev]{cleveref}

\usepackage{booktabs}       %
\usepackage{amsfonts}       %
\usepackage{microtype}      %
\usepackage{xcolor}         %
\usepackage{amsmath}
\usepackage{amssymb}
\usepackage{amsthm}
\usepackage{cleveref}
\usepackage{subcaption}
\usepackage{listings}
\usepackage[misc]{ifsym}

\theoremstyle{plain}

\theoremstyle{definition}

\theoremstyle{remark}

\usepackage[textsize=tiny]{todonotes}

\icmltitlerunning{MimicMotion: High-Quality Human Motion Video Generation with Confidence-aware Pose Guidance}

\begin{document}

\twocolumn[
\icmltitle{MimicMotion: High-Quality Human Motion Video Generation with Confidence-aware Pose Guidance}

\icmlsetsymbol{equal}{*}

\begin{icmlauthorlist}
\icmlauthor{Yuang Zhang}{tencent,sjtu}
\icmlauthor{Jiaxi Gu}{tencent}
\icmlauthor{Li-Wen Wang}{tencent}
\icmlauthor{Han Wang}{tencent,sjtu}
\icmlauthor{Junqi Cheng}{tencent}
\icmlauthor{Yuefeng Zhu}{tencent}
\icmlauthor{Fangyuan Zou}{tencent}
\end{icmlauthorlist}

\icmlaffiliation{tencent}{Tencent}
\icmlaffiliation{sjtu}{Shanghai Jiao Tong University}

\icmlcorrespondingauthor{Jiaxi Gu}{imjiaxi@gmail.com}

\icmlkeywords{Machine Learning, ICML}

\vskip 0.3in
]

\printAffiliationsAndNotice{}  %

\input{sections/teaser}

\input{sections/abstract}

\input{sections/introduction}

\input{sections/related_work}

\input{sections/method}

\input{sections/experiments}

\input{sections/conclusion}

\section*{Impact Statement}

MimicMotion advances human motion video generation technology with potential applications in creative content production, digital human, and entertainment. The framework enables the creation of high-quality, controllable human motion videos. However, like other video generation technologies, there are potential risks of misuse for creating deceptive content that could be used for misinformation. Responsible development and deployment of such technology should include appropriate safeguards, such as watermarking, detection methods, and clear usage guidelines. We encourage the development of technical capabilities and ethical frameworks for video generation technologies.

\bibliography{ms}
\bibliographystyle{icml2025}

\newpage
\appendix
\onecolumn
\section{Training Details}

MimicMotion is tuned from Stable Video Diffusion~\cite{blattmann_stable_nodate}. The video samples are resized to the resolution of 576 × 1024. The training dataset contains 4,436 human dancing videos collected from the internet. We train the model for 20 epochs at a batch size of 8. We follow Stable video diffusion and adopt the noise distribution, i.e.~$\log\sigma \sim \mathcal{N}(P_\text{mean}, P_\text{std}^2)$, proposed by Karras et al~\cite{karras_elucidating_2022} with parameter $P_\text{mean} = 0.5$ and $P_\text{std} = 1.4$. We train our model on 8 NVIDIA A100 GPUs with a batch size of 8 and 16 frames per clip. The loss weight assigned to the hand region is set at 10, while the remaining part is left to 1. The learning rate is $10^{-5}$ with a linear warmup schedule of 500 iterations. We tune all parameters in the UNet and PoseNet.

\section{PoseNet Model Architecture}

The PoseNet consists of convolutional and activation layers. The detailed architecture is listed in Table~\ref{tab:pose_net}.

\begin{table}[ht]
\centering
\small
\caption{Detailed Architecture of PoseNet}
\label{tab:pose_net}
\begin{tabular}{cccccc}
\toprule
\# & \textbf{Type} & \textbf{Channels} & \textbf{Kernel size} & \textbf{Stride} & \textbf{Pad} \\
\midrule
1  & Input         & 3                 & -                    & -               & -            \\
2  & Conv2d        & 3                 & 3x3                  & 1               & 1            \\
3  & SiLU          & -                 & -                    & -               & -            \\
4  & Conv2d        & 16                & 4x4                  & 2               & 1            \\
5  & SiLU          & -                 & -                    & -               & -            \\
6  & Conv2d        & 16                & 3x3                  & 1               & 1            \\
7  & SiLU          & -                 & -                    & -               & -            \\
8  & Conv2d        & 32                & 4x4                  & 2               & 1            \\
9  & SiLU          & -                 & -                    & -               & -            \\
10 & Conv2d        & 32                & 3x3                  & 1               & 1            \\
11 & SiLU          & -                 & -                    & -               & -            \\
12 & Conv2d        & 64                & 4x4                  & 2               & 1            \\
13 & SiLU          & -                 & -                    & -               & -            \\
14 & Conv2d        & 64                & 3x3                  & 1               & 1            \\
15 & SiLU          & -                 & -                    & -               & -            \\
16 & Conv2d        & 128               & 3x3                  & 1               & 1            \\
17 & SiLU          & -                 & -                    & -               & -            \\
18 & Conv2d        & 320               & 1x1                  & 1               & 0            \\
\bottomrule
\end{tabular}
\end{table}

\section{Additional Details on User Study}

\begin{figure}[h]
    \centering
    \includegraphics[width=0.7\linewidth]{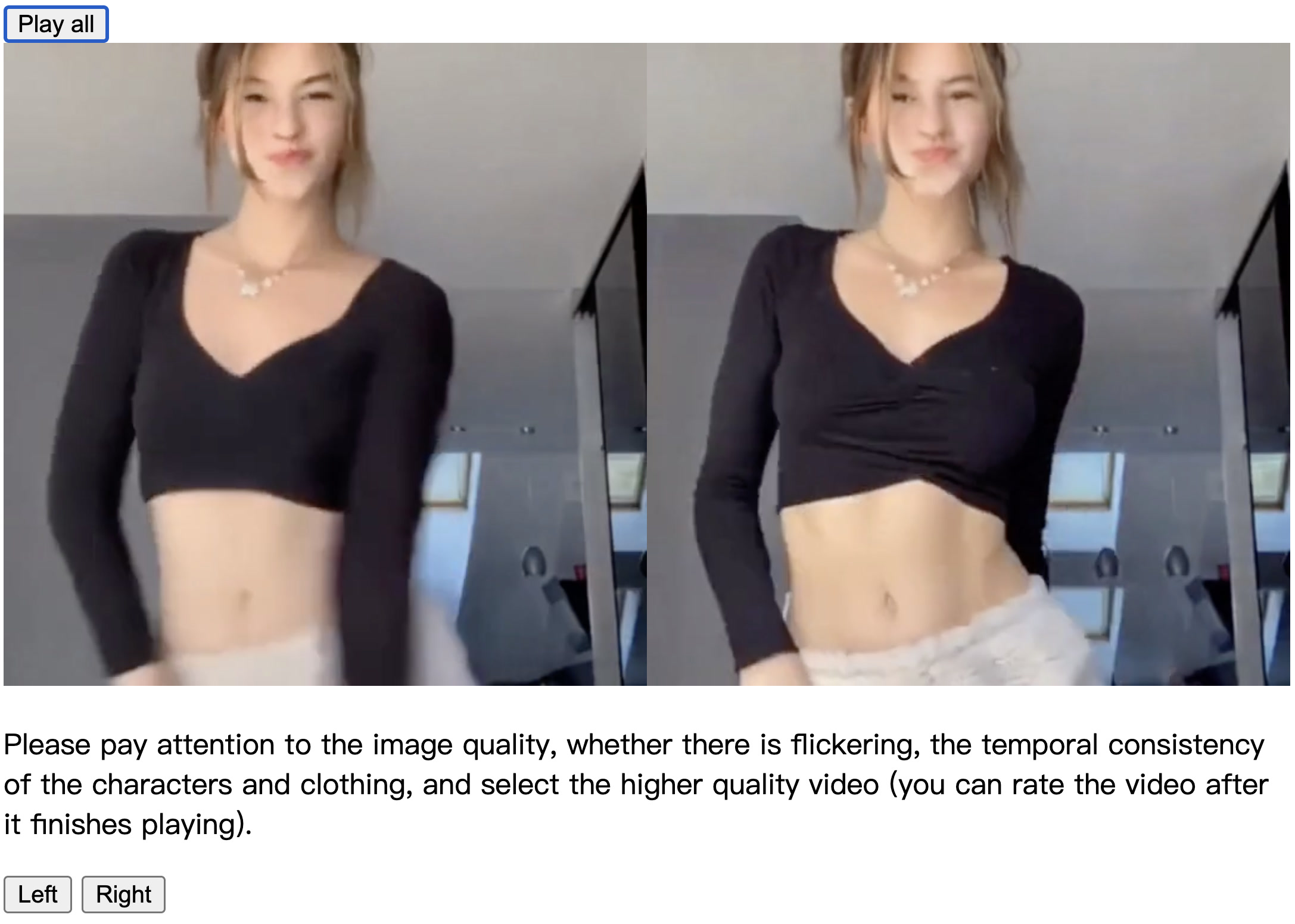}
    \caption{Snapshot of the user preference selection interface that prompts users to evaluate image quality and temporal smoothness. Note that the two videos play automatically, with our method randomly appearing on one side.}
    \label{fig:user_interface}
\end{figure}

The user study was designed to evaluate the subjective preferences of participants regarding the quality of videos generated by our method, MimicMotion, compared to several baseline methods on the TikTok dataset test split. Participants were instructed to select the video they perceived as having higher quality. The data collection process was conducted through an online interface, as depicted in Figure \ref{fig:user_interface}, which allowed participants to easily view and compare the video clips and submit their preferences. The results of our method will randomly appear in the video on the left or the right, with the other video being the comparative method.

\section{Cross-Domain Results: Cartoon and Animal Videos}

\begin{figure}
    \centering
    \includegraphics[width=\linewidth]{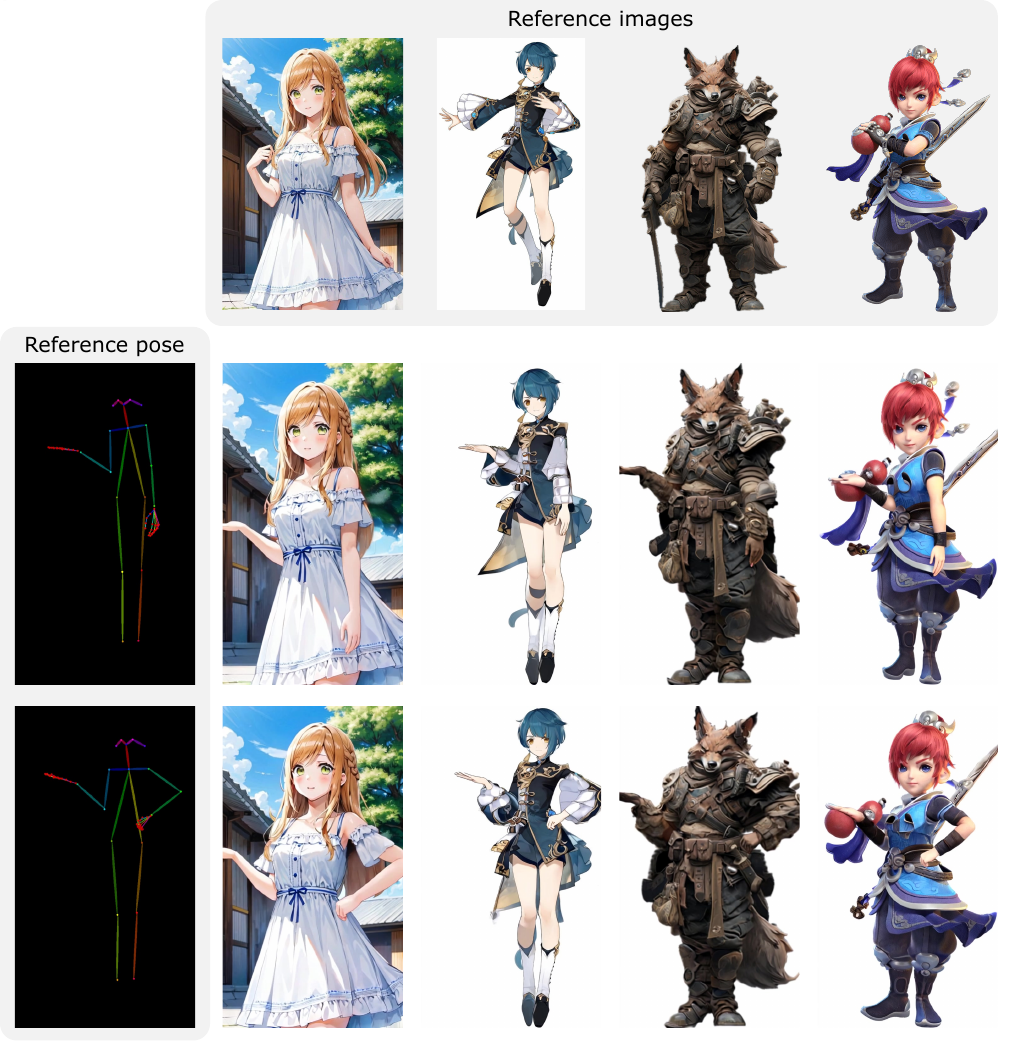}
    \caption{Generated cartoon dancing videos. Our method successfully preserves the artistic style of cartoon characters while generating natural motions that follow the input poses. Top: reference frame; Left: reference pose guidance; Bottom: generated frames at different timestamps.}
    \label{fig:cartoon}
\end{figure}

\begin{figure}
    \centering
    \includegraphics[width=\linewidth]{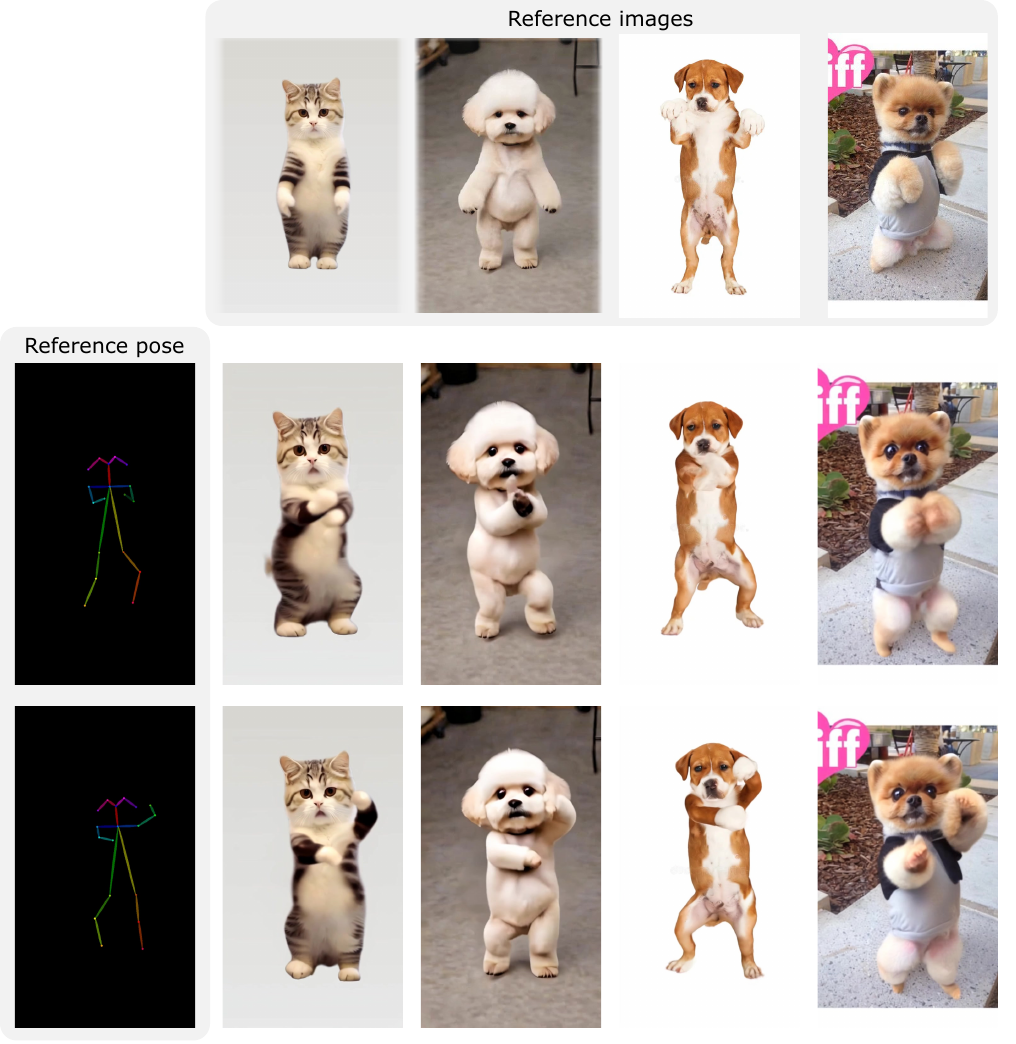}
    \caption{Generated animal dancing videos. MimicMotion can generalize to animal dancing zero-shot, producing plausible motions despite the significant appearance differences from humans. Top: reference frame; Left: reference pose guidance; Bottom: generated frames at different timestamps.}
    \label{fig:animal}
\end{figure}

To demonstrate the versatility of MimicMotion, we evaluate its performance on cross-domain scenarios, including 4 cartoon and 4 animal dancing videos. The videos are available on our anonymous project page. As shown in Figure \ref{fig:cartoon}, the model generates cartoon dancing videos while maintaining the artistic style of the characters. For animal subjects (Figure \ref{fig:animal}), our method produces reasonable results despite their significantly different appearance from humans.

These results highlight MimicMotion's generalizability, which stems from two key factors: first, the pose feature space of humans, cartoons, and animals may share a common subspace that is preserved in our training, human control signals can also guide cartoons and animals. Second, we map the limb lengths of the pose template to the reference character following MuseV~\cite{musev}, making the pose guidance signal closer to the reference character. This helps the model maintain the character's appearance while adapting poses across varying body proportions.

\end{document}

%% file: sections/teaser.tex
\begin{center}
  \captionsetup{type=figure}
  \includegraphics[width=\linewidth]{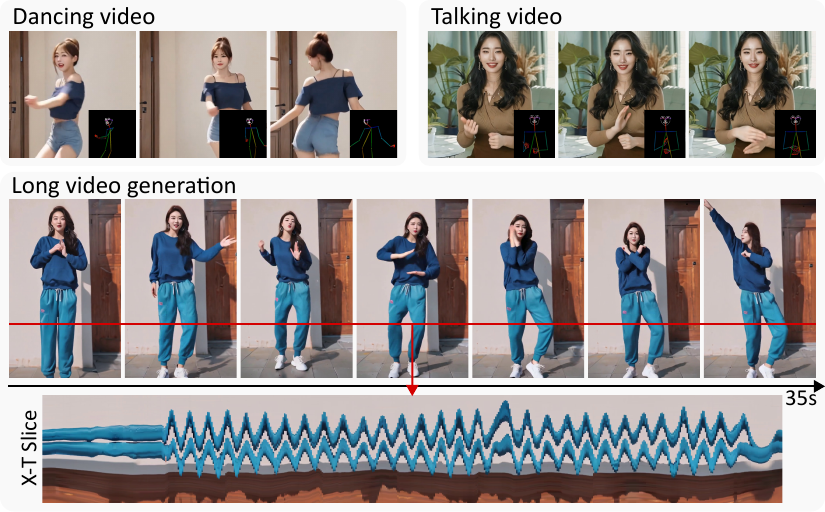}
  \captionof{figure}{Pose-guided dancing and talking videos generated by MimicMotion showcase its capability to produce diverse human motions and long videos.}
  \label{fig:teaser}
\end{center}

%% file: sections/abstract.tex
\begin{abstract}

In recent years, while generative AI has advanced significantly in image generation, video generation continues to face challenges in controllability, length, and detail quality, which hinder its application. We present \textit{MimicMotion}, a framework for generating high-quality human videos of arbitrary length using motion guidance.
Our approach has several highlights.
Firstly, we introduce confidence-aware pose guidance that ensures high frame quality and temporal smoothness. Secondly, we introduce regional loss amplification based on pose confidence, which reduces image distortion in key regions. Lastly, we propose a progressive latent fusion strategy to generate long and smooth videos. Experiments demonstrate the effectiveness of our approach in producing high-quality human motion videos. Videos and comparisons are available at \href{https://tencent.github.io/MimicMotion}{https://tencent.github.io/MimicMotion}.

\end{abstract}

%% file: sections/introduction.tex
\section{Introduction}
\label{sec:introduction}

Video generation has advanced alongside image generation. However, video generation is more challenging due to its higher inherent complexities, including the need for high-quality imagery and seamless temporal smoothness.
In addition, controlling the generated content and extension to significant lengths is essential for real-world use.
In this paper, we focus on pose-guided video generation with a reference image. Our goal is to generate a video that contains rich imagery details and adheres to the reference image and the pose guidance.

Currently, there are plenty of works focusing on this task, such as
DisCo~\cite{wang_disco_2024},
AnimateAnyone~\cite{hu_animate_2023},
MagicDance~\cite{chang_magicpose_2024},
MagicAnimate~\cite{xu_magicanimate_2023},
etc.
Though various techniques have been studied, the results are unsatisfactory in several aspects.
Imagery distortion especially on the regions of human hands is still a common issue which is particularly evident in videos containing large movements. Besides, to achieve good temporal smoothness, imagery details are sometimes sacrificed resulting in videos of blurred frames. In the presence of diverse appearances and motions in videos, accurate pose estimation is inherently challenging. This inaccuracy not only creates a conflict between pose alignment and temporal smoothness but also hinders the model scaling on the training schedule due to overfitting on noisy samples. In addition, due to computational limitations and model capabilities, there are still significant challenges in generating high-quality long videos containing a large number of frames. To solve these problems, we propose a series of approaches for generating long but still smooth videos based on pose guidance and image reference.

To alleviate the negative impact of inaccurate pose estimation, we propose an approach of confidence-aware pose guidance. By introducing the concept of confidence to the pose sequence representation, better temporal smoothness can be achieved and imagery distortion can also be eased. Confidence-based regional loss amplification can make the hand regions more accurate and clear. In addition, we propose a progressive latent fusion method for achieving long but still smooth video generation. Through generating video segments with overlapped frames with the proposed progressive latent fusion, our model can handle arbitrary-length pose sequence guidance. By merging the generated video segments, the final long video can be of good cross-frame smoothness and imagery richness at the same time (see Figure~\ref{fig:teaser}). For model training, to keep the cost of model training within an acceptable range, our method is based on a generally pre-trained video generation model. The amount of training data is not large and no special manual annotation is required.

In summary, there are three key contributions of this work:
\begin{enumerate}
\item We improve the pose guidance by employing a confidence-aware strategy. In this way, we alleviate the negative impact of inaccurate pose estimation. This approach not only reduces the influence of noisy samples during training but also corrects erroneous pose guidance during inference.
\item Based on the confidence-aware strategy, we propose hand region enhancement to alleviate hand distortion by strengthening the loss weight of the region of human hands with high pose confidence.
\item While overlapped diffusion is a standard technique for generating long videos, we advance it with position-aware progressive latent fusion that improves temporal smoothness at segment boundaries. Extensive experiments show the effectiveness of the proposed approach.
\end{enumerate}

%% file: sections/related_work.tex
\section{Related Work}
\label{sec:related_work}

\subsection{Diffusion Models for Image/Video Generation}

Diffusion-based models have demonstrated promising results in the fields of image~\cite{ho2020denoising, song2020score, ldm, controlnet, wang2024instantid} and video generation~\cite{makeavideo, guo_animatediff_2023, vdm, lvdm, videocrafter1, videocrafter2, blattmann_stable_nodate}, renowned for their capacity in generative tasks. Diffusion models operating in the pixel domain encounter challenges in generating high-resolution images due to information redundancy and high computational costs. Latent Diffusion Models (LDM)~\cite{ldm} address these issues by performing the diffusion process in low-dimensional latent spaces, significantly enhancing generation efficiency and quality while reducing computational demands. Recently, efforts have been made to improve training efficiency through noise schedule~\cite{zheng2024non,zheng2024beta}. Compared to image generation, video generation demands a more precise understanding of spatial relationships and temporal motion patterns. Recent video generation models leverage diffusion models by adding temporal layers to pre-trained image generation models~\cite{makeavideo,guo_animatediff_2023,tuneavideo,wang2024magicvideo}, or utilizing DiT~\cite{peebles2023scalable} structures to enhance generative capabilities for videos~\cite{yan2021videogpt,yu2023magvit,ma2024latte,bao2024vidu}. Stable Video Diffusion (SVD)~\cite{blattmann_stable_nodate} is one of the most popular open-source models built upon LDM. It offers a straightforward and effective method for image-based video generation and serves as a powerful pre-trained model for this task. Our approach extends SVD for pose-guided video generation, leveraging its pre-trained generative capabilities.

\subsection{Pose-Guided Human Motion Transfer}
Pose-to-appearance mapping aims to transfer motion from the source identity to the target identity. 
Methods based on paired keypoints from source and target images employ local affine transformations~\cite{FOMM, MRAA} or Thin-Plate Spline transformations~\cite{zhao2022thin} to warp the source image to match the driving image. These techniques aim to minimize distortion by applying weighted affine transformations, thereby generating poses in the output image that closely resemble those in the driving image.
Similarly, methods such as~\cite{chan2019everybody,tu2024stableanimator, ma_follow_2024, hu_animate_2023, feng_dreamoving_2023,peng2024controlnext,li2024dispose} utilize pose stick figures obtained from off-the-shelf human pose detectors as motion indicators and directly generate video frames through generative models. Depth information~\cite{feng_dreamoving_2023} or 3D human parametric models, such as SMPL (Skinned Multi-Person Linear)~\cite{zhu_champ_2024,wang2024vividpose}, can also be used to represent human geometry and motion characteristics from the source video.
Nevertheless, these overly dense guidance techniques can rely too much on the signal from the source video, such as the outline of the body, leading to a degradation in the quality of the generated videos, especially when the target identity differs significantly from the source.
Our approach, leveraging off-the-shelf human pose detectors, is capable of capturing the motion of the human body in driving videos without introducing excessive extraneous information, thereby ensuring the overall quality of the generated video.
In addition, we introduce confidence-aware pose guidance, which effectively mitigates the influence of inaccurate pose estimation in training and inference. In this way, we achieve superior portrait frame quality, especially in the hand regions.

\subsection{Long video generation}
Recent diffusion-based video generation algorithms are constrained to producing videos with durations of only a few seconds, significantly limiting their practical applications. As a result, substantial research efforts have been dedicated to extending the duration of generated videos, leading to the proposal of various approaches to overcome this limitation. Methods like~\cite{lvdm, voleti2022mcvd} autoregressively predict successive frames, enabling the generation of infinitely long videos. However, these methods often face quality degradation due to error accumulation and the lack of long-term temporal coherence. Hierarchical approach~\cite{yin2023nuwaxl, wang2024magicvideo} are proposed for generating long videos in a coarse-to-fine manner. It first creates a coarse storyline with keyframes using a global diffusion model, then iteratively refines the video with local diffusion models to produce intermediate frames.

MultiDiffusion~\cite{bar2023multidiffusion} combines multiple processes that use pre-trained text-to-image diffusion models to create high-quality images with user-defined controls. It works by applying the model to different parts of an image and using an optimization method to ensure all parts blend seamlessly. This allows users to generate images that meet specific requirements, like certain aspect ratios or spatial layouts, without needing additional training or fine-tuning. Lumiere~\cite{bar-tal_lumiere_nodate} extends MultiDiffusion to video generation by dividing the video into overlapping temporal segments. Each segment is independently denoised, and an optimization algorithm then combines these denoised segments. This approach ensures high coherence in the generated video, effectively maintaining temporal smoothness across segments. However, our experiments reveal that abrupt transitions can still occur at segment boundaries.

Building upon this principle, we introduce a position-aware progressive latent fusion strategy that enhances temporal smoothness near segment boundaries. We adaptively assign fusion weight based on the temporal position, ensuring a smooth transition at the segment boundaries.

%% file: sections/method.tex
\section{Method}
\label{sec:method}

\subsection{Preliminaries}

A Diffusion Model (DM) learns a diffusion process that generates a probability distribution for a given dataset.
In the case of visual content generation tasks, a neural network of DM is trained to reverse the process of adding noise to real data so new data can be progressively generated starting from random noise.
For a data sample $\mathbf{x} \sim p_{\text{data}}$ from a specific data distribution $p_{\text{data}}$, the forward diffusion process is defined as a fixed Markov Chain that gradually adds Gaussian noise to the data, following:
$
q(\mathbf{x_t} \mid \mathbf{x}_{t-1}) = \mathcal{N}(\mathbf{x}_t; \sqrt{1 - \beta_t} \mathbf{x}_{t-1}, \beta_t\mathbf{I})
$
for $t = 1, \cdots T$, where $T$ is the number of perturbing steps and $\mathbf{x}_t$ represents noisy data after adding $t$ steps of noise on the real data $\mathbf{x}_0$.
This process is controlled by a schedule $\beta_t$.
The above diffusion process can be reformed as:
\begin{equation}
    q(\mathbf{x_t} \mid \mathbf{x}_0) = \mathcal{N}(\mathbf{x}_t; \sqrt{\bar{\alpha}_t} \mathbf{x}_0, (1 - \bar{\alpha}_t)\mathbf{I})
\end{equation}
where $\bar{\alpha}_t=\prod_{i=1}^t \alpha_i$ and $\alpha_t = 1 - \beta_t$. Following DDPM~\cite{ho2020denoising}, a denoising function $\epsilon_\theta$ with parameter $\theta$ is trained by minimizing the mean squared error:
\begin{equation}
  \mathbb{E}_{\epsilon \sim \mathcal{N}(\mathbf{0}, \mathbf{I}), \mathbf{x}_t, \mathbf{c}, t} [\lVert \epsilon - \epsilon_{\theta} (\mathbf{x}_t; \mathbf{c}, t) \rVert_2^2]
\end{equation}
where $\mathbf{c}$ is the condition. We then train $\epsilon_\theta$ until convergence.

\subsection{Data Preparation}
\label{sec:dataprep}

To train a pose-guided video diffusion model, we collect a video dataset containing various human motions. The dataset need not be excessively large, as the pre-trained model already has a good prior.

Given a video from our dataset, the training sample is constructed with three parts: a reference image (denoted as $I_\text{ref}$), a sequence of raw video frames, and the corresponding poses. Basic pre-processing operations of resizing and cropping are applied to the raw video to get a sequence of video frames. For a given video, a fixed number of frames are randomly sampled at equal intervals as input video frames to the diffusion model.
The input reference image is independently sampled from the video and pre-processed in the same way as the video frames.
Another input of the model is the pose sequence, which is extracted from the video frames with DWPose~\cite{yang_effective_2023} frame by frame.

\subsection{Pose-Guided Video Diffusion Model}

\begin{figure}
    \centering
    \includegraphics[width=\linewidth]{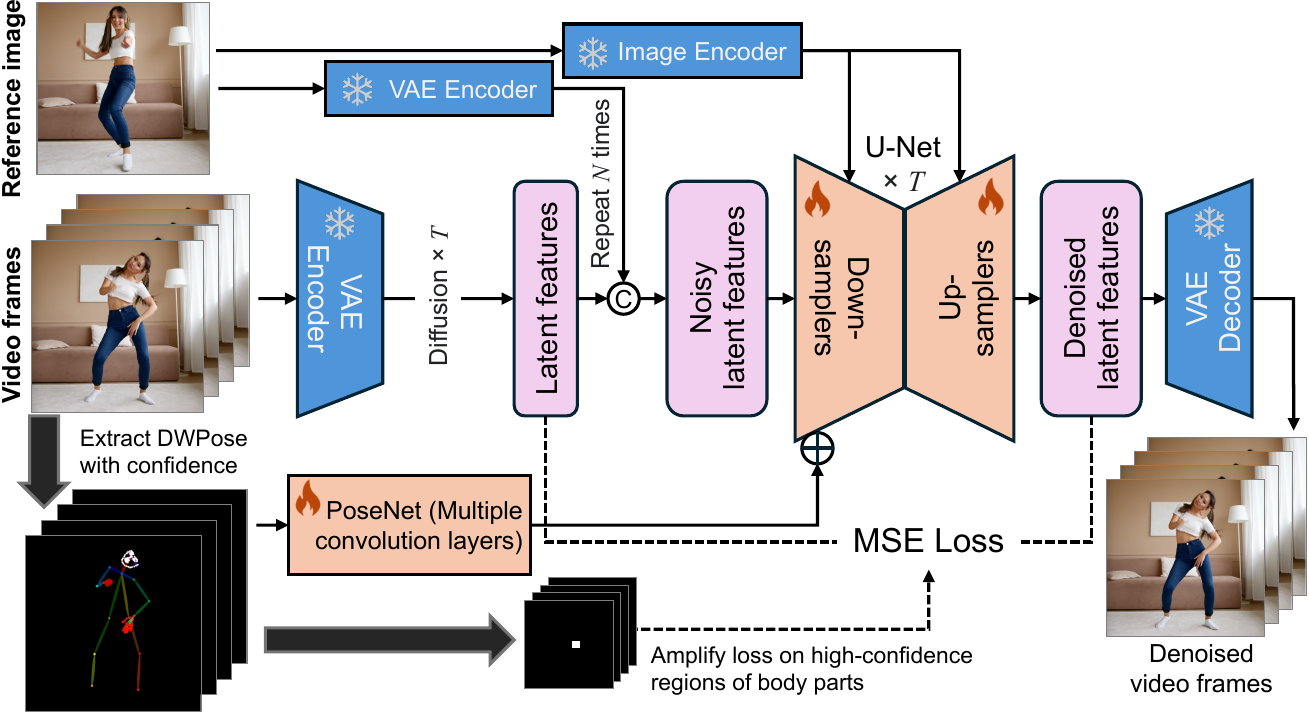}
    \caption{
MimicMotion integrates an image-to-video diffusion model with confidence-aware pose guidance, utilizing a spatiotemporal U-Net and a PoseNet for introducing pose sequence as the condition.
Our design features adaptive adjustment of pose influence via keypoint confidence scores and prioritizes high-confidence hand regions in the loss function.
}
\label{fig:model_structure}
\end{figure}

The goal of MimicMotion is to generate high-quality, pose-guided human videos from a single reference image and a sequence of poses to mimic.
This task involves the synthesis of realistic motion that adheres to the provided pose sequence while maintaining visual fidelity to the reference image.
We exploit the ability of a specific pre-trained video diffusion model to reduce the data requirement and computational cost of training a video diffusion model from scratch. Stable Video Diffusion (SVD)~\cite{blattmann_stable_nodate} is an open-source image-to-video diffusion model trained on a large-scale video dataset. It shows good performance on both video quality and diversity compared with the other contemporary models. The model structure of MimicMotion is designed to integrate a pre-trained Stable Video Diffusion (SVD) model to leverage its image-to-video generation capabilities.

Learning a diffusion process in pixel space is costly, especially for high-definition videos involving many frames. We follow the Latent Diffusion Model (LDM)~\cite{rombach2022high} to encode pixels into a lower-dimensional latent space. LDM adopts a pair of autoencoders, consisting of an encoder $\mathcal{E}$ and a decoder $\mathcal{D}$. Given a data sample $\mathbf{x}$, it is encoded into the latent space as $\mathbf{z} = \mathcal{E}(\mathbf{x})$. Conversely, the latent vector $\mathbf{z}$ can be decoded into pixel space via $\mathbf{x} = \mathcal{D}(\mathbf{z})$.

\Cref{fig:model_structure} shows the structure of our model. The VAE encoder on input video frames and the corresponding decoder for getting denoised video frames are both adopted from SVD and these parameters are frozen. The VAE encoder is applied independently to each frame of the input video and the conditional reference image. In contrast, the VAE decoder processes the latent features that undergo spatiotemporal interaction from U-Net. To enhance the temporal smoothness of low-level details, the VAE decoder incorporates temporal layers alongside the spatial layers.

The reference image and the sequence of poses are two other inputs of the model. The reference image follows two pathways: First, it is processed through CLIP~\cite{radford2021learning} to extract features that are fed into each U-Net block's cross-attention. Second, the image is encoded by the frozen VAE encoder to obtain its latent representation, which is then temporally duplicated to match the video frame dimensions. This latent representation is concatenated with the video frame features along the channel dimension before entering the U-Net.

For pose guidance, PoseNet consisting of multiple convolution layers is designed to extract features from input pose sequences. The reason for not using the VAE encoder is that the pixel value distribution of the pose sequence is different from that of common images on which the VAE autoencoder is trained. The features of poses are then element-wisely added to the output of the first convolution layer of U-Net.
Pose guidance is not added to every U-Net block due to: a) the sequence pose being extracted without temporal interaction may confuse the spatio-temporal layers within U-Net; b) excessive involvement of the pose sequence may degrade the performance of the pre-trained image-to-video model.

\subsection{Confidence-Aware Pose Guidance}

Inaccurate pose estimation has a negative impact on the model's training and inference.
However, accurate pose estimation is challenging in dynamic videos.
The limited capability of the pose estimation model, like DWPose~\cite{yang_effective_2023}, is only one aspect of the reason.
The more significant cause is the inherent uncertainty of pose from dynamic appearances and motions.
Specifically, incorrect pose guidance signals can mislead the model, resulting in the generation of inaccurate or distorted outputs, as illustrated in Figure~\ref{fig:posealpha}.
Moreover, noisy pose guidance signals can lead to overfitting on samples with incorrect poses, potentially causing training instability. This in turn may hinder the model's training.

For this problem, we propose confidence-aware pose guidance, which leverages the confidence scores associated with each keypoint from the pose estimation model. These scores reflect the likelihood of accurate detection, with higher values indicating higher visibility, less self-occlusion from other body parts, and motion blur.
Instead of applying a fixed confidence threshold to filter the keypoints, as commonly adopted in prior works~\cite{moore_anymate_2024,chang_magicpose_2024}, we utilize brightness of the keypoints and limb to represent the confidence level of pose estimation. Specifically, we multiply the color assigned to each keypoint and limb by its confidence score. Consequently, keypoints and corresponding limbs with higher confidence scores will appear more significant on the pose guidance map. This method enables the model to prioritize more reliable pose information in its guidance, thereby enhancing the overall accuracy of pose-guided generation.
In this way, the uncertainty of pose estimation can be conveyed through the pose guidance, making pose guidance more informative.

\paragraph{Hand region enhancement}
Moreover, we employ pose estimation and the associated confidence scores to alleviate region-specific artifacts, such as hand distortion, which are prevalent in the diffusion-based image and video generation models.
Specifically, we identify reliable regions via thresholding keypoint confidence scores. By setting a threshold, we can distinguish between keypoints that are confidently detected and those that may be unreliable due to self-occlusion or motion blur. Keypoints with confidence scores above the threshold are considered reliable.
We implement a masking strategy that generates masks based on a confidence threshold.
We unmask areas where confidence scores surpass a predefined threshold, thereby identifying reliable regions.
When computing the loss of the video diffusion model, the loss values corresponding to the unmasked regions are amplified by a certain scale so they can have more effect on the model training than other masked regions.

Specifically, to mitigate hand distortion, we compute masks using a confidence threshold for keypoints in the hand region. Only hands with all keypoint confidence scores exceeding this threshold are considered reliable, as a higher score correlates to higher visual quality. We then construct a bounding box around the hand by padding the boundary of these keypoints, and the enclosed rectangle is designated as unmasked. This region is subsequently assigned a larger weight in the loss calculation during the training of the video diffusion model. This selective unmasking and weighting process biases the model's learning towards hands, especially hands with higher visual quality, effectively reducing distortion and improving the overall quality.

\subsection{Progressive Latent Fusion for Long Videos}

\begin{figure}
    \centering
    \includegraphics[width=\linewidth]{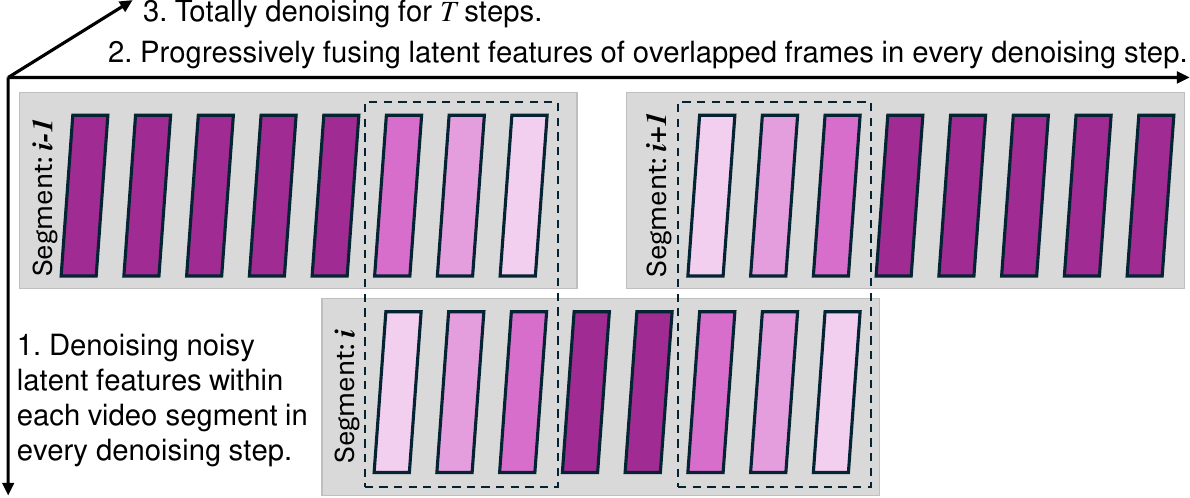}
    \caption{Overview of our approach for long video generation. The colored boxes represent latent video frames. A darker color means a higher weight. The dashed boxes represent video frame features involved in latent fusion.}
    \label{fig:latent_fusion}
\end{figure}

Limited by computation resources, generating long videos containing a large number of frames is challenging. 
For this problem, prior works like MultiDiffusion~\cite{bar2023multidiffusion} successfully use latent fusion for panoramic image generation. A similar idea can be applied to the video generation task. A straightforward approach is directly applying MultiDiffusion in the time domain, as in Lumiere~\cite{bar-tal_lumiere_nodate}. However, unlike spatial discontinuities between image tiles, temporal discontinuities are more noticeable to viewers, resulting in flickering or abrupt content changes. We address this by a progressive approach for generating long videos with high temporal continuity.

Progressive latent fusion is training-free and is integrated into the denoising process of the latent diffusion model during inference. \Cref{fig:latent_fusion} shows an overview of this process. We omit the VAE for brevity. The denoising process is done in latent space in our method. In general, there are $T$ denoising steps in total and our latent fusion is applied within each step. For a long given pose sequence, we use a pre-defined strategy for splitting the whole sequence into segments, consisting of a fixed number of frames per segment (denoted as $N$), with a certain number ($C$) of overlapped frames between every two adjacent segments. For the sake of generation efficiency, it is common to assume that $C \ll N$. During each denoising step, video segments are firstly denoised separately with the trained model, conditioning on the same reference image and the corresponding sub-sequence of poses.
Algorithm~\ref{alg:latent_fusion} shows the specific details of progressive latent fusion. As inputs, the reference image is denoted as $I_\text{ref}$, and the pose frame corresponding to $j$-th frame in $i$-th video segment is denoted as $P_i^j$. We use $\mathbf{z}_i^j$ to denote the latent feature of $j$-th frame in $i$-th video segment. The denoising process starts from the maximum time step $T$ and the latent features are initialized with a normal distribution $\mathcal{N}(\mathbf{0}, \mathbf{I})$. Within each denoising step at time step $t$, the reversed diffusion process defined by the trained model ($\text{DM}$) is applied to the latent features of each video segment numbered $i$ separately, with $\mathbf{z}_i$, $I_\text{ref}$, $P_i$ and $t$ as inputs. During the latent fusion stage, for every two adjacent video segments, the involved video frames are then fused. To avoid the corruption of temporal smoothness near video segment boundaries after latent fusion, we propose progressive latent fusion. For a video frame involved in latent fusion, its fusion weight is determined by its relative position in the video segment it belongs. Specifically, if a frame is close to the segment it belongs to, it will be assigned a heavier weight. For implementation, a fusion scale is pre-defined as $\lambda_\text{fusion}=1/(C+1)$ for controlling the level of latent fusion.

\begin{algorithm}[t]
\caption{Progressive latent fusion for long videos.}
\label{alg:latent_fusion}
\begin{algorithmic}[1]
\STATE {\bfseries Input:}
\STATE $I_\text{ref}$: Reference image;
\STATE $P_i^j$: Pose of $j$-th frame in $i$-th video segment;
\STATE $\mathbf{z}_i^j$: The latent of $j$-th frame in $i$-th video segment;
\STATE $N$: The number of frames in each video segment;
\STATE $C$: The number of overlapped frames.
\STATE {\bfseries Output:} A long sequence of latent video frames.
\vspace{1ex}
\STATE $\mathbf{z} \sim \mathcal{N}(\mathbf{0}, \mathbf{I})$
\STATE $\lambda_\text{fusion} \gets 1/(C + 1)$

\FOR{$t=T$ {\bfseries to} $1$}
    \FOR{$i=1,2,\dots$}
        \STATE $\mathbf{z}_i \gets \text{DM}(\mathbf{z}_i, I_\text{ref}, P_i, t)$
    \ENDFOR
    \FOR{$i = 1,2,\dots$}
        \FOR{$j=1$ {\bfseries to} $N$}
            \IF{$i > 1$ \AND $j \leq C$}
                \STATE $\mathbf{z}_i^j \gets j \lambda_\text{fusion} \mathbf{z}_i^j + (1 - j\lambda_\text{fusion}) \mathbf{z}_{i-1}^{N-C+j}$
            \ELSIF{$j > N-C$}
                \STATE $\mathbf{z}_i^j \gets (N+1-j)\lambda_\text{fusion} \mathbf{z}_i^j + (C-N+j)\lambda_\text{fusion} \mathbf{z}_{i+1}^{C-N+j}$
            \ENDIF
        \ENDFOR
    \ENDFOR
\ENDFOR

\STATE {\bfseries return} $\{\mathbf{z}_i^{1:N-C}\}_{i=1,2,\cdots}$
\end{algorithmic}
\end{algorithm}

After applying $T$ iterative denoising steps, we aggregate the final long sequence by combining the denoised overlapped video segments in latent space. The resulting aggregated long video in latent space is then fed to the temporal VAE decoder to reconstruct the long video.

%% file: sections/experiments.tex
\section{Experiments}
\label{sec:experiments}

\subsection{Implementation Details}
Due to the lack of suitable high-quality academic datasets, we collect 4,436 human dancing videos from the internet for training.
The average length is 20.1s. We adopt the pre-trained weights from SVD~\cite{blattmann_stable_nodate}.
The PoseNet is trained from scratch.
We train our model on 8 NVIDIA A100 GPUs for 20 epochs, with a batch size of 8 and 16 frames per clip. The loss weight of the hand region is 10. The learning rate is $10^{-5}$ with a linear warmup of 500 iterations. We tune all parameters in the UNet and PoseNet.

\subsection{Comparison to State-of-the-Art Methods}

We compare our method with latest state-of-the-art pose-guided human video generation methods, including MagicAnymate~\cite{xu_magicanimate_2023}, MagicPose~\cite{chang_magicpose_2024}, Moore-AnymateAnyone~\cite{moore_anymate_2024}, and MuseV~\cite{musev}.
For the testing protocol of previous works~\cite{wang_disco_2024,chang_magicpose_2024}, we adopt the TikTok~\cite{Jafarian_2021_CVPR_TikTok} dataset and use sequence 335 to 340 for our evaluation.

We conduct qualitative and quantitative comparisons, complemented by a user study.
Each method has a different input aspect ratio. To ensure a fair comparison, we only consider the central square region of the videos.
Specifically, we individually apply a center crop to the reference image and pose sequence to match each method's input aspect ratio.
Then, we extract the center squares from the generated videos for comparison across different methods.

\begin{figure}
\centering
\includegraphics[width=\linewidth]{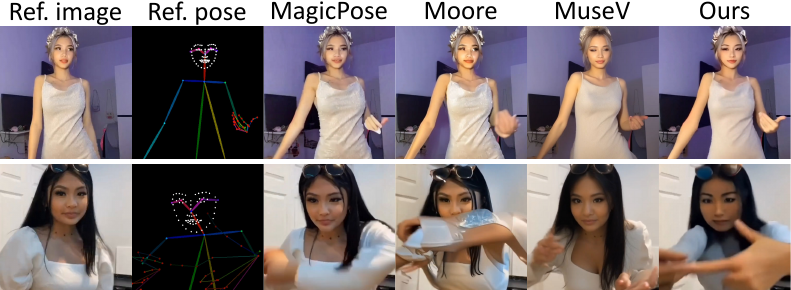}
\caption{Qualitative comparison to the state-of-the-art methods on TikTok test set. Our method achieves better hand generation quality and adheres better to the reference pose. Note that our method is not trained on the TikTok dataset.}
\label{fig:cmp}
\end{figure}

\begin{figure}
\centering
\includegraphics[width=\linewidth]{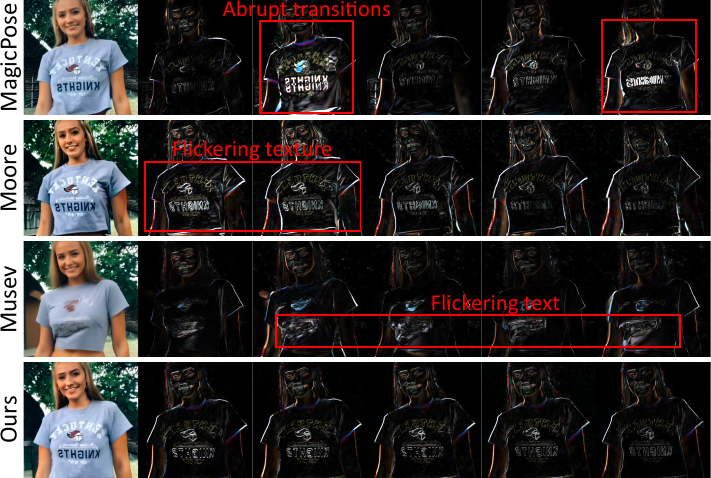}
\caption{Comparison of temporal smoothness via frame differences.
MagicPose exhibits abrupt transitions, whereas Moore and MuseV show flickering texture and text. In contrast, our method maintains stable frame differences.}
\label{fig:cmp_tm}
\end{figure}

\paragraph{Qualitative evaluation} We conduct qualitative comparisons between the selected baselines and our method. In \Cref{fig:cmp}, we showcase sample frames to highlight the superior quality of individual frames produced by our method. Additionally, in \Cref{fig:cmp_tm}, we illustrate the temporal differences, demonstrating the enhanced temporal smoothness of our approach compared to existing methods.

\Cref{fig:cmp} presents a comparison of the generated frames, where each row represents a distinct example. The first row demonstrates the superior hand quality achieved by our approach, while the second row showcases the improved adherence to pose guidance. These improvements directly result from our confidence-aware pose guidance and hand region enhancement design.

Importantly, our method shows superior temporal smoothness, characterized by smooth motion and minimal flickering. Figure~\ref{fig:cmp_tm} demonstrates this through pixel-wise differences between consecutive frames. While MagicPose~\cite{chang_magicpose_2024} shows abrupt transitions, Moore-AnymateAnyone~\cite{moore_anymate_2024} exhibits unstable clothing textures, and MuseV~\cite{musev} struggles with consistent text rendering, our method maintains stable frame-to-frame transitions without visible artifacts. Videos in the supplementary material further demonstrate this improvement.
This enhancement is enabled by our confidence-aware pose guidance, which mitigates the impact of inaccurate pose inputs and the associated temporal noise. By weighting pose signals by confidence, our method ensures high temporal smoothness in the presence of pose guidance noise.

\paragraph{Quantitative evaluation}
Our method outperforms existing approaches across all evaluation metrics, as shown in Table~\ref{tab:sota}. We evaluated performance on test sequences from the TikTok dataset~\cite{Jafarian_2021_CVPR_TikTok} using both video-level metrics FID-VID~\cite{balaji2019conditional} and FVD~\cite{unterthiner2018towards}, as well as frame-level metrics PSNR and SSIM~\cite{wang2004image} following~\cite{wang_disco_2024}. The results demonstrate our method's superior performance compared to current state-of-the-art approaches.

\begin{table}[]
\small
\centering
\setlength{\tabcolsep}{5pt}
\begin{tabular}{@{}l|cccc@{}}
\toprule
Method & FID-VID$\downarrow$ & FVD$\downarrow$ & SSIM$\uparrow$ & PSNR$\uparrow$ \\ \midrule
MagicAnymate & 16.2    &     848 & 0.740 & 17.5\\
MagicPose    & 13.3    &     916 & 0.776 & 18.8\\
Moore        & 12.4    &     728 & 0.758 & 18.7\\
MuseV        & 14.6    &     754 & 0.766 & 17.6\\
MimicMotion (ours) & \bf 9.3 & \bf 594 & \bf 0.795 & \bf 20.1\\
\bottomrule
\end{tabular}
\caption{Quantitative comparison to state-of-the-art methods: MagicAnymate~\cite{xu_magicanimate_2023},
MagicPose~\cite{chang_magicpose_2024},
Moore~\cite{moore_anymate_2024},
MuseV~\cite{musev}.
We evaluate on the TikTok dataset test split. MimicMotion is the best in all metrics.}%
\label{tab:sota}
\end{table}

\begin{figure}[t]
\centering
  \includegraphics[width=\linewidth]{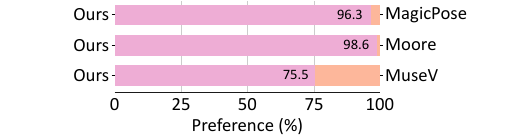}
  \caption{Preference of videos generated by MimicMotion (ours) over baseline methods on the TikTok dataset test split. Users consistently prefer MimicMotion over other methods.}
  \label{fig:user_study}
\end{figure}

\paragraph{User study}
To supplement our quantitative and qualitative evaluations, we conduct a user study to assess the subjective preferences of participants regarding the generated videos on the TikTok dataset test split.
The study involves showing two video clips—one generated by our method and the other by one of the baseline methods—to a diverse group of users.
Participants are instructed to select the video that they perceived as having higher quality, considering factors including image quality and temporal smoothness of characters and clothing.
We collected data from 36 participants, with each participant evaluating 6 video pairs for our method against each baseline method. As shown in Figure~\ref{fig:user_study}, the results indicate a strong preference for MimicMotion over the baseline methods. In comparison to MagicPose and Moore, the participants almost favored all videos produced by our method. Despite MuseV showing higher image quality compared to other baselines, the preference for videos produced by our method still reached 75.5\%.
These findings align with our qualitative and quantitative evaluation, reinforcing the effectiveness of our method in meeting user expectations for high-quality human video generation.

\subsection{Ablation Study}

As shown in \Cref{tab:ablation_results}, confidence-aware pose guiding and hand region enhancement improve all metrics. Progressive latent fusion notably improves FVD scores, indicating better temporal coherence, though other metrics may not reflect this improvement. We analyze each component below.

\begin{table}
\centering
\setlength{\tabcolsep}{4pt}
\small
\begin{tabular}{ccc|cccc}
\toprule
Hand & Conf. & Prog. & FID-VID$\downarrow$ & FVD$\downarrow$ & SSIM$\uparrow$ & PSNR$\uparrow$ \\
\midrule
  &   &   & 14.6 & 776 & 0.760 & 18.0 \\
  &   & $\checkmark$ & 15.0 & 678 & 0.758 & 17.9 \\
  & $\checkmark$ & $\checkmark$ & 12.2 & 623 & 0.787 & 18.4 \\
$\checkmark$ & $\checkmark$ & $\checkmark$ & \bf 9.3 & \bf 594 & \bf 0.795 & \bf 20.1 \\
\bottomrule
\end{tabular}
\caption{Ablation studies of hand region augmentation (hand), confidence-aware pose guiding (conf.), and progressive latent fusion (prog.).}
\label{tab:ablation_results}
\end{table}

\paragraph{Confidence-aware pose guiding}

\begin{figure}[]
\centering
\includegraphics[width=\linewidth]{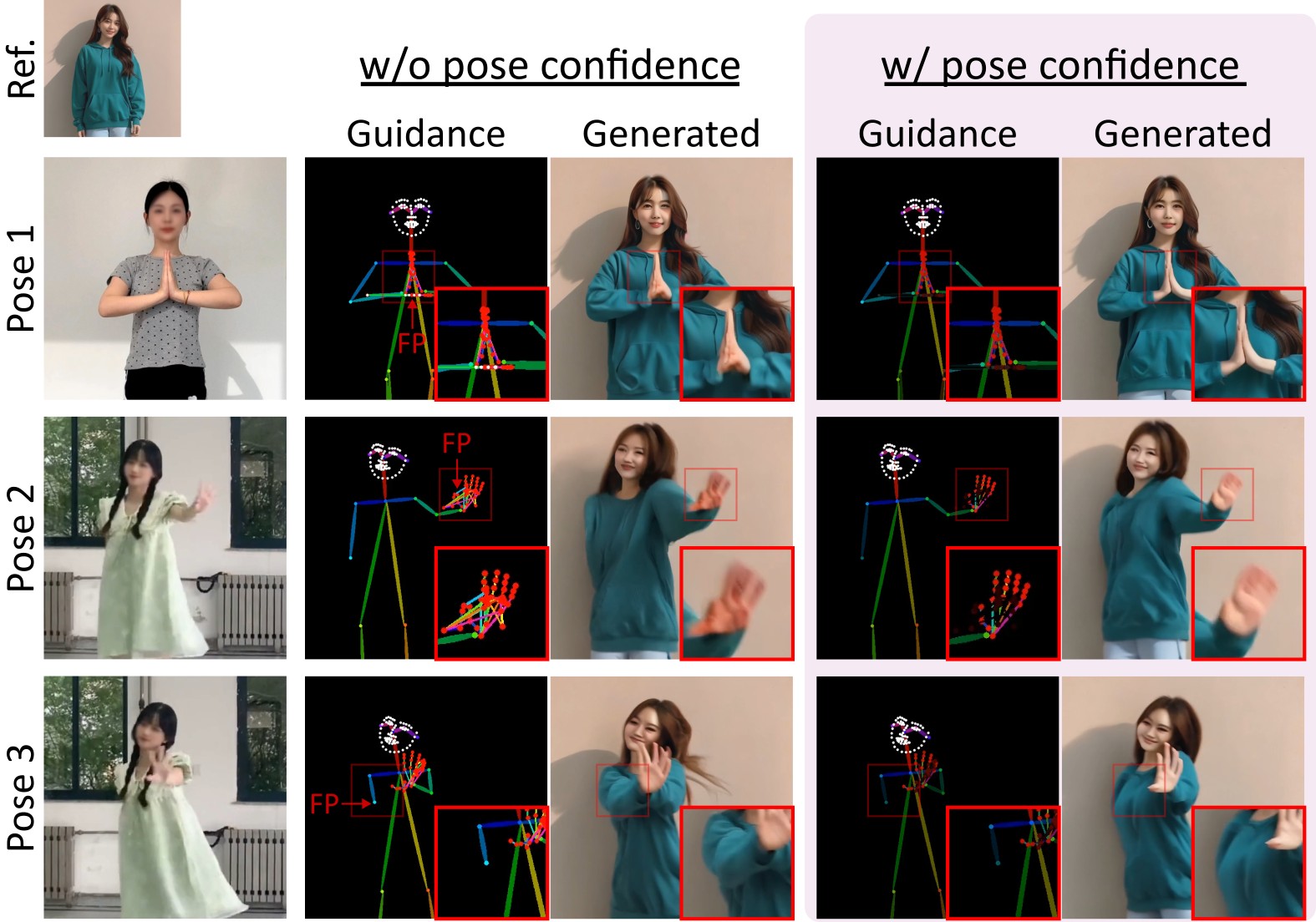}
\caption{Confidence-aware pose guiding enhances generation robustness to false guiding signals (Pose 1\&2) and provides hints to handle self-occlusion (Pose 3).}
\label{fig:posealpha}
\end{figure}

Figure~\ref{fig:posealpha} shows the effectiveness of confidence-aware pose guidance. Each row corresponds to one sample. The left side shows three images to extract the pose. On the right side, we plot the corresponding pose guiding signals, both with and without confidence-aware pose guiding. From the guiding signals, we can see that there are errors in the pose estimated by DWPose. Nevertheless, our confidence-aware design minimizes the impact of incorrect pose estimation in guidance signals.

Specifically, in the case of Pose 1, the estimation exhibits a duplicate detection issue, which leads to the inclusion of duplicate keypoints. In the case of Pose 2, there is one hand obscured, and the keypoints of this hand are incorrectly estimated on the other hand; In the case of Pose 3, the right elbow is obscured, but it is still detected with confidence above the threshold thus falsely remains in the guidance signal. These problems lead to confusing hand guidance signals and ultimately lead to distortions such as deformed hands or wrong spatial relationships in the generated frames.

In contrast, by integrating confidence scores into pose guidance, our method effectively mitigates these issues. The confidence scores provide a measure of reliability for each keypoint, allowing our model to weigh the guidance signals accordingly. Specifically, keypoint with lower confidence, which typically correspond to inaccurate keypoints caused by self-occlusion or motion blur, will be of less significance in the guidance. This approach leads to clearer and richer pose guidance, as the influence of potentially erroneous keypoints is reduced. The corresponding generation results demonstrate how our method enhances the robustness of generation against false guiding signals (Pose 1 and Pose 2) and offers visibility hints to resolve the front-back ambiguity of 2D pose estimation (Pose 3).

\paragraph{Hand region enhancement}

\begin{figure}[]
\centering
\includegraphics[width=\linewidth]{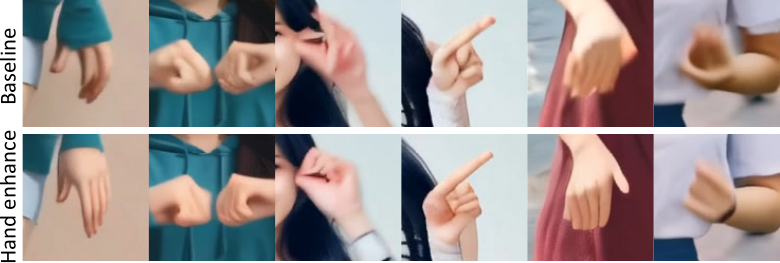}
\caption{Hand region enhancement consistently reduces hand distortion and improves visual appeal, using the same reference image and pose guidance.}
\label{fig:handaug}
\end{figure}

In conjunction with confidence-aware pose guiding, we further improve hand generation quality by assigning a higher weight to the hand region in the training loss. Figure~\ref{fig:handaug} compares the generation result with and without hand region enhancement, using the same reference image and pose guidance. All experiments incorporate confidence-aware pose guidance. The hands in the first row are cropped from the generated video frames of a model trained without hand region enhancement, which exhibits noticeable distortions, such as irregular and misplaced fingers. In contrast, the results of the model trained with hand region enhancement (the second row) show consistent improvements in hand generation quality and a reduction in hand distortion.
These results show the effectiveness of the proposed hand region enhancement design, which substantially mitigates hand distortion, which is a prevalent challenge in diffusion-based models.
As humans often focus on hands, this design aligns the training process with human preferences, thereby enhancing the overall visual quality of the results. There is no noticeable degradation in the quality of the remaining parts.

\paragraph{Progressive latent fusion}

\begin{figure}[]
\centering
\begin{subfigure}{\linewidth}
    \centering
    \includegraphics[width=\linewidth]{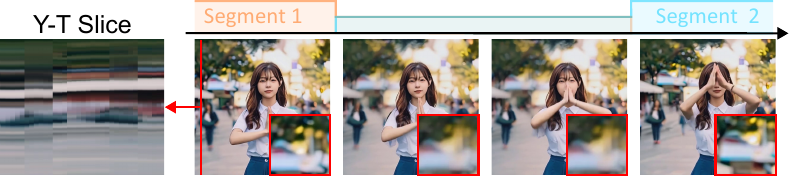}
    \caption{Without progressive latent fusion.}
    \label{fig:multidiffA}
\end{subfigure}
\begin{subfigure}{\linewidth}
    \centering
    \includegraphics[width=\linewidth]{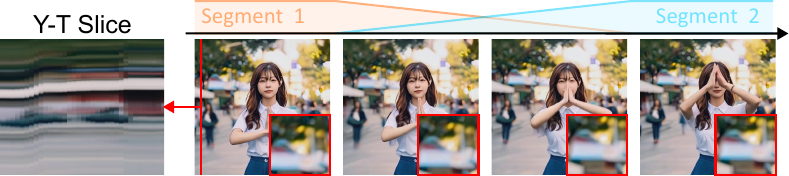}
    \caption{With progressive latent fusion.}
    \label{fig:multidiffB}
\end{subfigure}
\caption{Progressive latent fusion ensures smooth segment transitions by avoiding temporal discontinuities of weights in simple averaging, which reduces artifacts in Y-T slices.}
\label{fig:multidiff}
\end{figure}

To achieve seamless transitions between video segments, we introduce progressive latent fusion, a technique that gradually blends frames in the overlapped regions of consecutive video segments.
Figure~\ref{fig:multidiff} presents a comparison of a sample transition with an overlap of 6 frames between two 16-frame segments. We visualize 4 sample frames.
The original MultiDiffusion approach employs a simple averaging of frames within the overlap region. This method assigns equal weight to all frames in the overlap region, irrespective of their temporal position (whether they are closer to the preceding or subsequent segment). It lacks a gradual transition from one segment to another, causing noticeable flickers at denoising window boundaries.
This is evident in the y-t slice shown on the left of Figure~\ref{fig:multidiffA}, where the transition at segment boundaries is abrupt. The right side of the figure shows four frames at the segment boundary. Note that the background in the top-left corner (enlarged) is initially clear in segment 1. It suddenly becomes blurry in the overlapped region and then suddenly reverts to clearer in the main part of segment 2. This artifact is not observed when progressive latent fusion is applied.

The proposed progressive latent fusion approach (see Figure~\ref{fig:multidiffB}) mitigates these issues. The left y-t slice shows smooth transitions across segment boundaries, avoiding abrupt changes. The right side of the figure demonstrates that this approach eliminates a sudden blurring, mitigating flicking artifacts, thus improving the overall visual temporal coherence for long video generation.

%% file: sections/conclusion.tex
\section{Conclusion}
\label{sec:conclusion}

In this study, we introduce MimicMotion, a pose-guided human video generation model that leverages confidence-aware pose guidance and progressive latent fusion for producing high-quality, long videos with human motion guided by pose.
Through extensive experiments and ablation studies, we show that our model achieves superior adaptation to noisy pose estimation, enhancing hand quality and ensuring temporal smoothness. The integration of confidence scores into pose guidance, the enhancement of hand region loss, and the implementation of progressive latent fusion are crucial in achieving these improvements, resulting in more visually compelling and realistic human video generation.